\documentclass[runningheads]{llncs}
\usepackage{cite}
\usepackage{graphicx}

\usepackage{amsmath,amsthm,amssymb,amsfonts}
\usepackage{booktabs,multirow}
\usepackage{algorithm}
\usepackage{algpseudocode}
\usepackage{bbding}
\usepackage{float}
\usepackage[cmyk]{xcolor}
\usepackage{cite}
\usepackage{color}
\usepackage[colorlinks,
            linkcolor=blue,
            anchorcolor=blue,
            citecolor=blue]{hyperref}
\usepackage[numbers,sort&compress]{}
\begin{document}
\bibliographystyle{splncs04}
\title{DP-DCAN: Differentially Private Deep Contrastive Autoencoder Network for Single-cell Clustering}
%
%
\author{Huifa Li\inst{1}\inst{\star} \and
Jie Fu\inst{1}\thanks{: These authors contributed equally.} \and
Zhili Chen\inst{1} \inst{(}\Envelope\inst{)} \and
Xiaomin Yang\inst{2} \inst{(}\Envelope\inst{)} \and
Haitao Liu\inst{1} \and
Xinpeng Ling\inst{1}
}

\institute{Shanghai Key Laboratory of Trustworthy Computing, East China Normal University, Shanghai, China \\
\email{\{huifali,jie.fu,htliu,xpling\}@stu.ecnu.edu.cn}, \email{zhlchen@sei.ecnu.edu.cn} \and
Xinhua Hospital, School of Medicine, Shanghai Jiao Tong University, Shanghai, China \\
\email{yangxiaomin@xinhuamed.com.cn}}

\authorrunning{H. Li et al.}
%
%
%
\maketitle              
\begin{abstract}
Single-cell RNA sequencing (scRNA-seq) is important to transcriptomic analysis of gene expression. Recently, deep learning has facilitated the analysis of high-dimensional single-cell data. Unfortunately, deep learning models may leak sensitive information about users. As a result, Differential Privacy (DP) is increasingly being used to protect privacy. However, existing DP methods usually perturb whole neural networks to achieve differential privacy, and hence result in great performance overheads. To address this challenge, in this paper, we take advantage of the uniqueness of the autoencoder that it outputs only the dimension-reduced vector in the middle of the network, and design a {\underline D}ifferentially {\underline P}rivate {\underline D}eep {\underline C}ontrastive {\underline A}utoencoder {\underline N}etwork (DP-DCAN) by partial network perturbation for single-cell clustering. Firstly, we use contrastive learning to enhance the feature extraction of the autoencoder. And then, since only partial network is added with noise, the performance improvement is obvious and twofold: one part of network is trained with less noise due to a bigger privacy budget, and the other part is trained without any noise. Experimental results of 8 datasets have verified that DP-DCAN is superior to the traditional DP scheme with whole network perturbation.

\end{abstract}
\section{Introduction}
Single-cell RNA sequencing (scRNA-seq) has enabled unbiased, high-throughput, and high-resolution transcriptome analysis at the level of individual cells \cite{yu2023topological,flores2022deep}, playing a crucial role in identifying cell types through transcriptome analysis, investigating developmental biology, uncovering intricate diseases and predicting cell developmental pathways.
Therefore, the accurate identification of cell types is a crucial step in scRNA-seq analysis. Clustering is a very powerful method for cell type annotation, as it can identify cell types in an unbiased manner.
However, the quality of feature representation has remained a bottleneck for the performance of clustering.
Besides, scRNA-seq data associated with diseases are highly sensitive and need rigorous privacy protection~\cite{bonomi2020privacy}.

In general, deep autoencoder is a type of neural network with the primary purpose of learning an informative representation of the data that can be used for different applications and has emerged as a powerful tool for clustering. 
In this work, we focus on scRNA-seq data clustering. In the setting of scRNA-seq data clustering, the dataset consists of multiple cells, and the autoencoder is trained to learn a useful representation for each cell. The representation is input to a classifier to predict a specific cell type. 
In recent years, many novel deep learning clustering approaches with autoencoder have successfully developed to model the high-dimensional and sparse scRNA-seq data, such as scDC \cite{tian2019clustering}, scDCC \cite{tian2021model}, bmVAE \cite{yan2023bmvae}, DCA \cite{eraslan2019single}, scSemiCluster \cite{chen2021single} and scVAE \cite{gronbech2020scvae}.

However, drawing meaningful insights in many of these approaches fundamentally relies upon the utilization of privacy-sensitive, often scarce, training data associated with individuals. scRNA-seq data yields some important privacy concerns, including genetic conditions \cite{ocasio2019scrna} and predispositions to specific diseases \cite{oestreich2021privacy}.
In contrast to the modifiability of compromised credit card numbers, scRNA-seq data, once revealed, remain unchangeable. The revelation of these holds the potential to detrimentally affect patients by influencing their prospects in employment and education.


It is thus apparent that the implementation of privacy-enhancing techniques is required to facilitate the training of models of sensitive scRNA-seq data. One of the state-of-the-art paradigms to prevent privacy disclosure in machine learning is differential privacy (DP)~\cite{abadi2016deep,fu2022adap,fu2023dpsur,mironov2017renyi}.
DP is a pioneering approach in the domain of privacy-preserving data analysis. It operates on the principle of introducing controlled noise into the data or its analysis results, which is carefully calibrated to protect individual privacy without significantly compromising the data's utility for group-level inferences.
Several studies have demonstrated that the implementation of differential privacy (DP) can effectively mitigate the inadvertent disclosure of private training data, including vulnerabilities to membership inference attacks \cite{leemann2024gaussian,ye2022one,houssiau2022difficulty} during the training process.
Many applications of differential privacy in the fields of biology and medicine have been proposed \cite{chen2020differential,tramer2015differential}.
However, there is no differentially private learning methods have taken scRNA-seq data into account.


Unfortunately, the random noise introduced by differential privacy leads to poor performance of current differential privacy deep learning models \cite{bagdasaryan2019differential}. Additionally, research indicates that the noise introduced by differential privacy increases as the model size grows \cite{shen2021towards}. Achieving good performance of trained models while ensuring differential privacy protection is a challenge, especially in autoencoders, as they are more sensitive to random noise \cite{ha2019differential}.

In this paper, we first introduce a {\underline D}ifferentially {\underline P}rivate {\underline D}eep {\underline C}ontrastive {\underline A}utoencoder {\underline N}etwork for single-cell dimension reduction and clustering (DP-DCAN).
The algorithm leverages autoencoders to learn discriminative features of instances and utilizes contrastive learning to enhance the extraction of features by contrasting the similarity and dissimilarity of samples under different clusters. In addition, we incorporate differential privacy into the process to protect the privacy of clustering outputs. However, adding differential privacy protection at the autoencoder stage is non-trivial because the random noise introduced by differential privacy affects the training effectiveness of the autoencoder. Therefore, based on the post-processing property of differential privacy \cite{zhu2021bias}, we designed a differential privacy method with partial network perturbation that can effectively reduce the impact of noise on our autoencoder model. The experimental results demonstrated that our algorithm DP-DCAN can achieve excellent performance under differential privacy protection on real scRNA-seq datasets.

The main contributions of our work are summarized below:

\begin{itemize}
    \item We proposed a DP-DCAN algorithm. To the best of our knowledge, this is the first article to propose the differential privacy framework for single-cell clustering.
    \item We designed a differential privacy method with the partially perturbed network for autoencoder and the method can efficiently mitigate the performance sacrifice of the autoencoder model.
    \item We conducted a rigorous privacy analysis on the proposed method based on Rényi differential privacy, proving that it meets differential privacy.
    \item Comprehensive empirical evaluation on 8 public datasets. The results confirm that DP-DCAN can achieve superior clustering performance under the protection of differential privacy.
\end{itemize}

The rest of the paper is organized as follows. Section \ref{sec:related} provides an overview of the related work. Section \ref{sec:preli} introduces the preliminary knowledge. In Section \ref{sec:methods}, we present the details of our method. Privacy analysis is conducted in Section \ref{sec:privacy_ana} and the experimental results are presented in Section \ref{sec:results}. Finally, Section \ref{sec:conclusion} is the conclusion. 

\section{Related Work} \label{sec:related}
\subsection{Single-cell RNA-seq Data Clustering}
Clustering is an unsupervised method capable of accurately identifying single-cell types. Deep embedding clustering methods perform clustering and feature learning which can address the challenge of high heterogeneity and sparsity in scRNA-seq data.

In the past years, deep learning methods have been used and advanced to analyze scRNA-seq data. Eraslan et al. \cite{eraslan2019single} introduce the ZINB-based autoencoder to capture the complexity and non-linearity in scRNA-seq data. In another work, Tian et al. \cite{tian2019clustering} leverage a deep autoencoder as an embedding method, which simultaneously learns feature representation and clustering via explicit modeling of scRNA-seq data generation. Tian et al. \cite{tian2021model} convert prior knowledge into soft pairwise constraints and add them as additional terms into the loss function for optimization. Moreover, Wang et al. \cite{wang2021scgnn} and Yu et al. \cite{yu2022zinb} aggregate cell-cell relationships and identify cell clusters based on deep graph convolutional networks.

\subsection{Differential Privacy in Medicine and Bioinformatics}
There are a few works that take into account scRNA-seq data in differentially private learning.
Liu et al. \cite{liu2019adaptive} propose a definition of adaptive differential privacy of character and its mechanisms satisfying expected privacy-preserving and expected data utility.
Wu et al. \cite{wu2019differential} makes use of the closed frequent pattern set to reduce redundant motifs of result sets and obtain accurate motifs results, satisfying $\epsilon$-differential privacy.
Islam et al. \cite{islam2022differential} propose a differential privacy-based DL framework to solve biological problems: breast cancer status (BCS) and cancer type (CT) classification, and drug sensitivity prediction.
Yilmaz et al. \cite{yilmaz2022genomic} propose a mechanism that considers the correlations in data during data sharing, eliminates statistically unlikely data values beforehand, and adjusts the probability distributions for each shared data point accordingly.
To the best of our knowledge, the problem of how to achieve differentially private learning with the scRNA-seq data has not been well-addressed.

\section{Preliminaries} \label{sec:preli}

In this section, we introduce and formalize the theory to train deep autoencoder using the concept of differentially private stochastic gradient descent (DP-SGD).

\subsection{Deep Autoencoder}
This semi-supervised learning method can be implemented by deep autoencoder(DAE) \cite{rumelhart1986learning}. The DAE is typically composed of two main components: an encoder network and a decoder network, corresponding to $f_e$ and $f_d$.

In particular, the autoencoder learns a map from the input to itself through a pair of encoding and decoding phases:
\begin{align}
    \hat{X} &= f_d(f_e(X)).
\end{align}
where $X$ is the input data, $f_e$ is an encoding map from the input layer to the hidden layer, $f_d$ is a decoding map from the hidden layer to the output layer, and $\hat{X}$ is the reconstruction of the input data. The idea is to train $f_e$ and $f_d$ to minimize the difference between $X$ and $\hat{X}$.

\subsection{Differential Privacy}
Differential privacy is a mathematically rigorous framework that formally defines data privacy. It requires that a single entry in the input dataset cannot result in statistically significant changes in the output \cite{dwork2006calibrating, dwork2011firm, dwork2014algorithmic} if differential privacy holds.
\begin{definition}
\textit{(($\epsilon, \delta$)-Differential Pricacy)}. The randomized mechanism $A$ provides ($\epsilon$,  $\delta$)-Differential Privacy (DP), if for any two neighboring datasets $D$ and $D'$ that differ in only a single entry, $\forall$S $\subseteq$ Range($A$),
\end{definition}
\begin{equation}
{\rm Pr}(A(D) \in S) < e^{\epsilon} \times {\rm Pr}(A(D') \in S) + \delta.
\end{equation}
Here, $\epsilon > 0$ controls the level of privacy guarantee in the worst case. The smaller $\epsilon$ is, the stronger the privacy level is. The factor $\delta > 0$ is the failure probability that the property does not hold. In practice, the value of $\delta$ should be negligible~\cite{zhu2017differential,papernot2018scalable}, particularly less than $\frac{1}{|D|}$.

\subsection{Differentially Private Stochastic Gradient Descent}

Abadi et al.\cite{abadi2016deep} proposed Differentially Private Stochastic Gradient Descent (DP-SGD) which
is a widely adopted scheme for training deep neural networks with differential privacy guarantees. Specifically, in each iteration $t$, a batch of tuples $\mathcal{B}_t$ is sampled from $D$ with a fixed probability $\frac{b}{|D|}$, where $b$ is batch size. After computing the gradient of each tuple $x_i \in \mathcal{B}_t$ as $g_t(x_i) = \nabla_{\theta_i} L(\theta_i,x_i)$, where $\theta_i$ is model parameter for the $i$-th sample, DPSGD clips each per-sample gradient according to a fixed $\ell_{2}$ norm ~\eqref{eq:clipping}.

\begin{align}
	\begin{split}
		\overline{g}_t\left(x_{i}\right) 
		& = \text{Clip}(g_t\left(x_{i}\right);C), \label{eq:clipping} \\
		& = g_t\left(x_{i}\right) \Big/ \max \Big(1, \frac{\left\|g_t\left(x_{i}\right)\right\|_{2}}{C}\Big).
	\end{split}
\end{align}

In this way, for any two neighboring datasets, the sensitivity of the query $\sum_{i\in \mathcal{B}_t} g(x_i)$ is bounded by  $C$. Then, it adds Gaussian noise scaling with this norm to the sum of the gradients when computing the batch-averaged gradients:
\begin{equation}\label{eq:add noise}
\tilde{g}_t = \frac{1}{b}\left(\sum_{i \in \mathcal{B}_t} \overline{g}_t\left(x_{i}\right)+\mathcal{N}\left(0, \sigma^{2} C^{2} \mathbf{I}\right)\right),
\end{equation}
where $\sigma$ is the noise scale depending on the privacy budget. Last, the gradient descent is performed based on the batch-averaged gradients. Since initial models are randomly generated and independent of the sample data, and the batch-averaged gradients satisfy the differential privacy, the resulting models also satisfy the differential privacy due to the post-processing property. 

\subsection{Privacy Threat Model}
\begin{figure}[htbp]
	\centering
	\includegraphics[width=0.80\linewidth]{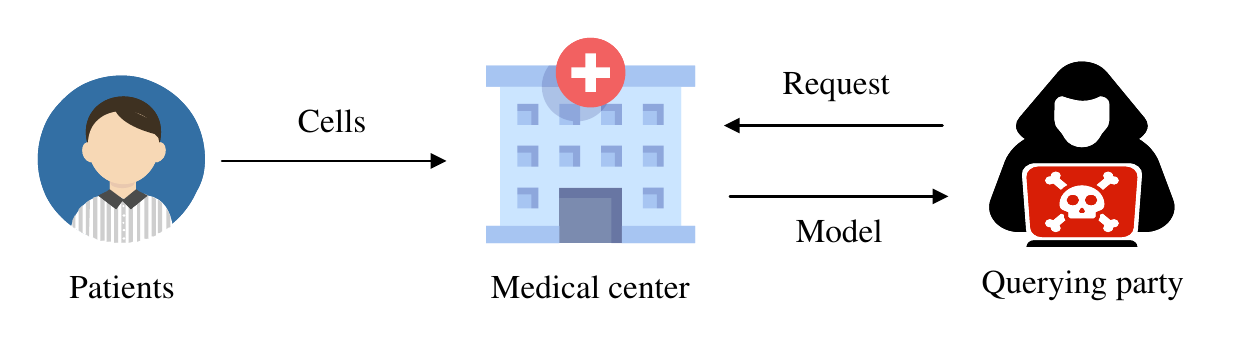}
    \caption{Privacy threat model of scRNA-seq data deep learning.}
    \label{fig:threat}
\end{figure}

In the context of deep learning models for scRNA-seq data sharing as shown in Fig. \ref{fig:threat}, it is presumed that the patients and medical center are entities considered trustworthy. Conversely, the querying party, which utilizes the shared model, is assumed to be honest but curious. Therefore, the querying party is curious about sensitive information in scRNA-seq data analysis. Also, considering the querying party has full background knowledge of genome data.
Patients send biological samples to the medical center, where scRNA-seq data can be obtained by sequencing. Therefore, the scRNA-seq data of the medical center is reliable. In the scRNA-seq data sharing model, the specific privacy threat model is as follows: The querying party is honest but curious; the querying party can obtain all background knowledge about patients' scRNA-seq data; Patients, medical center, and querying party are rational.

\section{Methodology} \label{sec:methods}
In this section, we first introduce the proposed method termed DP-DCAN. We then present our idea of a differentially private single-cell autoencoder network. And we introduce a contrastive learning module. Finally, we elaborate on the proposed loss function in DP-DCAN.

\begin{figure*}[htbp]
	\centering
	\includegraphics[width=0.95\linewidth]{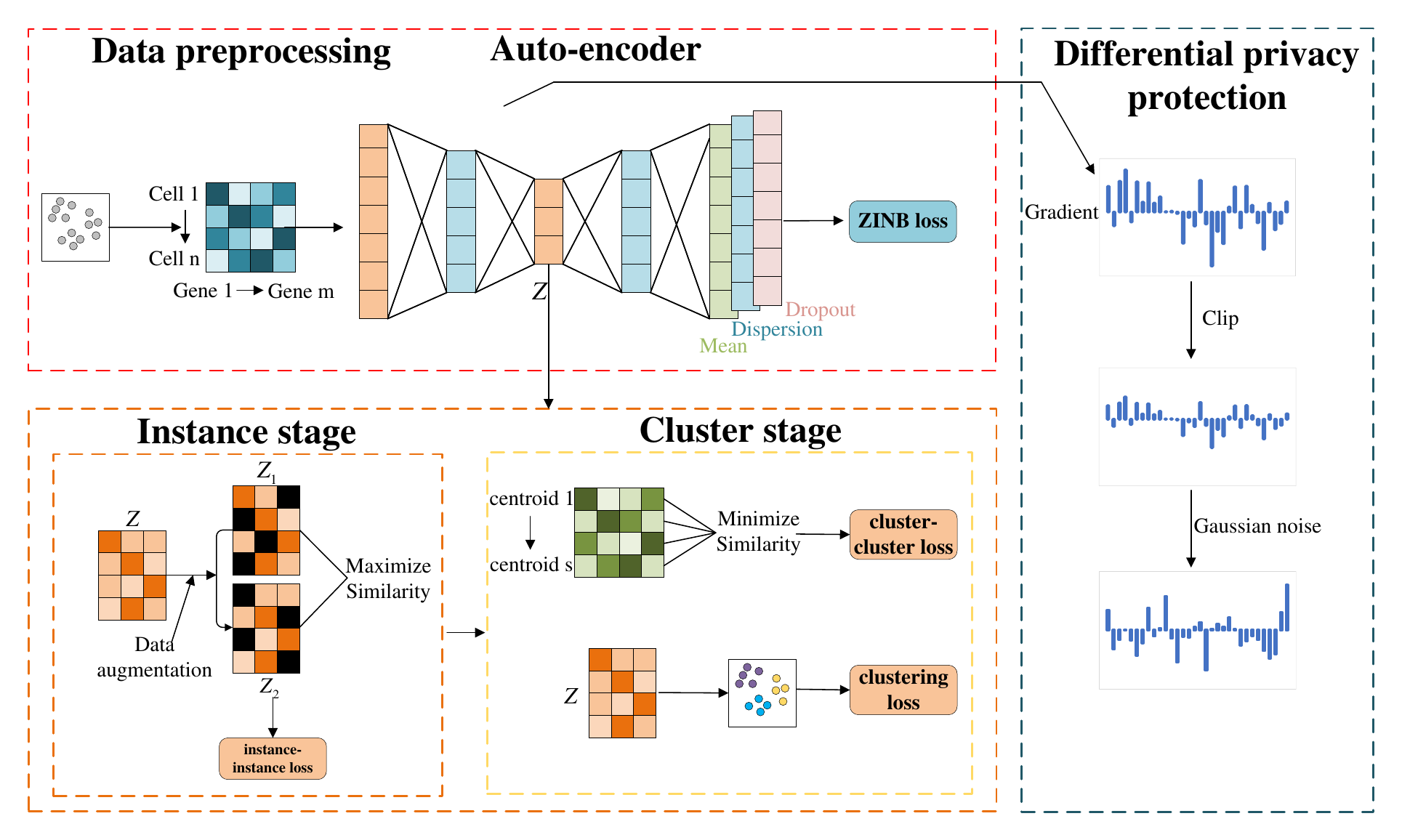}
    \caption{The overview of DP-DCAN.}
    \label{fig:framework}
\end{figure*}

\subsection{Overview}
As shown in Fig. \ref{fig:framework}, our DP-DCAN framework mainly consists of two key components: the private single-cell autoencoder network and the two-stage contrastive learning module. 
In the private single-cell autoencoder network, the autoencoder is trained to learn the essential feature representation of the scRNA-seq data while preventing leakage of sensitive information elegantly. In the two-stage contrastive learning module, we further enhance the autoencoder’s understanding of feature representation by comparing the similarities and disparities between different samples and across various clusters.

The detailed procedure of DP-DCAN is outlined in Algorithm \ref{code:dp-scc}. The algorithm employs a two-stage training. In the first stage, the autoencoder maximizes the similarity between instances from the same cluster in compressed feature space as shown in Lines \ref{code:inslearn1}-\ref{code:inslearn2}. The second stage enables the autoencoder to maximize the distance between different clusters as shown in Lines \ref{code:clslearn1}-\ref{code:clslearn2}. The design of loss functions $\mathcal{L}_{instance}$ and $\mathcal{L}_{cluster}$ for two-stage training is shown in Section \ref{sec:loss}. In the two stages, Gaussian noise is added to the encoder network $\theta_e$ of autoencoder while optimizing the model in Lines \ref{code:update1} and \ref{code:update2}. The decoder network $\theta_{d}$ of autoencoder doesn't be disturbed because it's useless during inference. Lines \ref{code:pl1} and \ref{code:pl2} keep track of the privacy loss of RDP and convert it to $(\epsilon,\delta)$ in Line \ref{code:pl4}.

\subsection{Differentially Private Single-cell Autoencoder Network} \label{subsec:pri_net}
The trained encoder can map the high-dimensional to low-dimensional and the trained decoder doesn't work during inference. Moreover, the existing DP methods usually perturb whole neural networks to achieve differential privacy and hence result in great performance overheads. Inspired by these, We proposed a differentially private deep autoencoder network that only adds noise to the gradient of the encoder while minimizing the empirical loss function for single-cell clustering. The network is based on a Zero-Inflated Negative Binomial (ZINB) loss function that can learn the hidden feature representation of scRNA-seq data. Within this framework, fully connected layers stacked in both the encoder and decoder facilitate the capture of intricate representation embeddings from scRNA-seq datasets.

\begin{algorithm}[htbp]
    \caption{DP-DCAN framework}
    \label{code:dp-scc}
    \begin{algorithmic}[1]
        \Require
        {Examples $X = \{x_1, x_2, \dots, x_N\}$, learning rate $\eta$, lot size $L$, noise scale $\sigma$, clipping bound $C$.}
        \Ensure {$\theta_e$ and privacy budget ($\epsilon, \delta$).}
        \State {Initialize autoencoder parameter $\theta = \{\theta_{e}, \theta_{d}\}$ randomly, $R_1$=0, $R_2=0$}
        \State{\textcolor{gray}{\# Instance stage}}
        \State{Design $\mathcal{L}_{instance}$ as shown in Equ. (\ref{eq:instance})}
        \For {$t_1 \in [1, T_1]$} \label{code:inslearn1}
            \State{$DPAN_{input} \leftarrow (\theta_{e}^{t_1}, \theta_{d}^{t_1}, \mathcal{L}_{instance}, X, L, C, \sigma,\eta)$}
            \State{$(\theta_{e}^{t_1+1}, \theta_{d}^{t_1+1}), R_1 \leftarrow DPAN(DPAN_{input})$} \label{code:update1}
             \State{$R_1 += R_1$} \label{code:pl1}
        \EndFor \label{code:inslearn2}
        \State{\textcolor{gray}{\# Cluster stage}}
        \State{Design $\mathcal{L}_{cluster}$ as shown in Equ. (\ref{eq:cluster})}
        \For {$t_2 \in [T_1+1, T_2]$} \label{code:clslearn1}
            \State{$DPAN_{input} \leftarrow (\theta_{e}^{t_{2}}, \theta_{d}^{t_2}, \mathcal{L}_{cluster}, X, L, C, \sigma,\eta)$}
            \State {$(\theta_{e}^{t_2+1}, \theta_{d}^{t_2+1}), R_2 \leftarrow DPAN(DPAN_{input})$} \label{code:update2}
            \State{$R_2 += R_2$} \label{code:pl2}
        \EndFor \label{code:clslearn2}
        \\
        Compute privacy budget $\epsilon$ by Theorem~\ref{the:privacy-loss-scCDP} \label{code:pl4}
    \\
    \Return {$\theta_{e}^{T_{2}}$ and $(\epsilon,\delta)$.}
    \end{algorithmic}
\end{algorithm}

Algorithm \ref{pdp} outlines our training approach with parameters $\theta$ by minimizing the empirical loss function $\mathcal{L}(\theta)$. The encoder parameters $\theta_e$ are perturbed when optimizing. And the decoder parameters $\theta_{d}$ don't be perturbed. At each optimization step, we compute the gradient $g(x_i)$ including protected gradient $g_{e}(x_i)$ and exposed parameters $g_{d}(x_i)$ for a random subset of samples, clip the $\ell_2$ norm of each gradient and compute the average in Lines \ref{code:sample}-\ref{code:clip2}. Then, we add noise to the gradient of $\theta_e$ in order to protect privacy with the smallest possible performance loss and take a step in the opposite direction of this average noisy gradient in Lines \ref{code:addnoise}-\ref{code:sgd_update}. At the end, we compute the privacy budget by Theorem~\ref{the:rdp-of-dpsgd} in Line \ref{code:rdp}. The privacy analysis is in Section \ref{sec:privacy_ana}.

\begin{algorithm}[htbp]
    \caption{Differentially private autoencoder network (DPAN)}
    \label{pdp}
    \begin{algorithmic}[1]
        \Require
        {Examples {$x_1, x_2, \dots, x_N$}, learning rate $\eta$, lot size $L$, loss function $\mathcal{L}(\theta)$, noise scale $\sigma$, clipping bound $C$.}
        \Ensure {$(\hat{\theta}_{e}, \hat{\theta}_{d})$ and privacy budget $RDP$.}
        \State {Take a random sample $\mathcal{B}$ with probability L/N}\label{code:sample}
        \For {$x_i \in \mathcal{B}$}
            \State {Compute $g_{e}(x_i),g_{d}(x_i) \leftarrow \nabla_{\theta}\mathcal{L}(\theta,x_i)$}
            \State $\bar{g_e}(x_i) \leftarrow g_e(x_i) / \mathrm{max}(1, \frac{\|g_{e}(x_i)\|_2}{C})$ \label{code:clip1}
            \State $\bar{g_d}(x_i) \leftarrow g_d(x_i) / \mathrm{max}(1, \frac{\|g_{d}(x_i)\|_2}{C})$ \label{code:clip2}
        \EndFor
        \State {$\tilde{g}_{e} \leftarrow \frac{1}{L} (\sum_i^L\bar{g}_{e}(x_i) + \mathcal{N}(0, \sigma^2C^2\textbf{I}))$} \label{code:addnoise}
        \State {$\tilde{g}_{d} \leftarrow \frac{1}{L} \sum_i^L\bar{g}_{d}(x_i)$}
        \State {$\hat{\theta}_{e} \leftarrow \theta_{e} - \eta \tilde{g}_{e}$}
        \State {$\hat{\theta}_{d} \leftarrow \theta_{d} - \eta \tilde{g}_{d}$} \label{code:sgd_update}
        \State Compute privacy budget $RDP$ by Theorem~\ref{the:rdp-of-dpsgd} \label{code:rdp}
        \\
    \Return {$(\hat{\theta}_{e}, \hat{\theta}_{d})$, $RDP$.}
    \end{algorithmic}
\end{algorithm}

\subsection{Contrastive Learning Module} \label{subsec:contra_lm}

To capture the relationships between cells, We integrated a two-stage contrastive learning module into our model to enhance the autoencoder's ability to discern more discriminative features within instances without negative pairs and clusters. In the first stage, specific samples $x$ are augmented into $\{x_1,x_2\}$ as positive examples. Then, the augmented data is mapped to a lower-dimensional space using the encoder. Inspired by SimSiam \cite{chen2021exploring}, which does not rely on negative pairs. We maximize the similarity between two augmentations of $x$ in order to converge towards improved performance. This approach learns more robust and useful representations by enhancing the similarity of samples from the same class. Instance-level contrastive learning extracts distinctive features by minimizing differences between these instances. We minimize their negative cosine similarity:
\begin{align}
    \mathcal{L}_{ii} &= -\frac{z_1}{||z_1||_2} \cdot \frac{z_2}{||z_2||_2} \label{eq:lii}.
\end{align}
where $z_1$, $z_2 $ are $x_1, x_2$ which are mapped to a low dimensional vector through an encoder, and $||\cdot||$ is $\ell_2$-norm.

In the second stage, based on the centroids as representative positions for each cluster, cluster-level contrastive learning is applied to capture discriminative cluster features. Utilizing K-means, we acquire centroid positions. Cluster-level contrastive learning is committed to maximizing the distance between clusters, ensuring clusters are positioned as distantly from each other as feasible:
\begin{align}
    \mathcal{L}_{cc} &= \frac{1}{s^2}\sum_{i=0}^{s} \sum_{j=0}^{s} \frac{c_i}{||c_i||_2} \cdot \frac{c_j}{||c_j||_2} \label{eq:lcc}.
\end{align}
where $c_i$ is the centroid of cluster $i$, $c_j$ is the centroid of cluster $j$ and $s$ is the number of clusters.
\subsection{Joint Embedding and Clustering Optimization} \label{sec:loss}
We use a hybrid loss to optimize our proposed DP-DCAN during the two-stage training phase. Different from the two-stage training proposed by Wang\cite{wang2023scdcca} that can't compute the gradient for each individual sample in a batch which is not suitable for differential privacy, our method is designed to calculate the gradient by an individual sample. Specifically, in the first stage, we jointly combined the reconstruction loss and the instance-instance loss to minimize the distance of the instances of the same cluster. Reconstruction loss $\mathcal{L}_{zinb}$, instance-instance loss $\mathcal{L}_{ii}$ and hybrid loss $\mathcal{L}_{instance}$ are defined as follows:
\begin{align}
    \mathrm{ZINB}(x|\pi,\mu,\theta) &= \pi\delta_0(x) + (1-\pi)\mathrm{NB}(x|\mu,\theta), \\
    \mathrm{NB}(x|\mu,\theta) &= \frac{\Gamma(x+\theta)}{x!\Gamma(\theta)} (\frac{\theta}{\theta+\mu})^{\theta} (\frac{\mu}{\theta+\mu})^{x}, \\
    \mathcal{L}_{zinb}(X, \hat{X}) &= -\mathrm{log}(\mathrm{ZINB}(x|\pi,\mu,\theta)), \label{eq:lzinb} \\
    \mathcal{L}_{instance} &= \rho \mathcal{L}_{zinb} + (1-\rho) \mathcal{L}_{ii}. \label{eq:instance}
\end{align}

where $x$ is the expression value in cell, $\pi$ is the probability of a dropout event, $\mu$ is the mean, and $\theta$ represents the divergence of the negative binomial distribution. $\rho$ is a weighted factor that balance the impact of $\mathcal{L}_{zinb}$ and $\mathcal{L}_{ii}$.

In the second stage, we employ reconstruction loss, clustering loss, and cluster-cluster loss to maximize the distance between clusters. Clustering loss represents the disparity between the clustering results and the true data distribution which is measured by Kullback–Leibler (KL) divergence \cite{kullback1951information}. Clustering loss $\mathcal{L}_{cls}$, cluster-cluster loss $\mathcal{L}_{cc}$ and hybrid loss $\mathcal{L}_{cluster}$ are defined as follows:
\begin{align}
    q_{ij} &= \frac{\left(1+\left\|z_i-\lambda_j\right\|^2\right)^{-1}}{\sum_{j'}\left(1+\left\|z_i-\lambda_{j'}\right\|^2\right)^{-1}}, \\
    p_{ij} &= \frac{q_{ij}^2/\sum_jq_{ij}}{\sum_{j^{\prime}}\left(q_{ij^{\prime\prime}}^2/\sum_{j^{\prime}}q_{ij^{\prime}}\right)}, \\
    L_{cls} &= KL\left(P||Q\right)=\sum_i\sum_jp_{ij}\log\frac{p_{ij}}{q_{ij}}, \\
    \mathcal{L}_{cluster} &= \beta_1 \mathcal{L}_{zinb} + \beta_2 \mathcal{L}_{cls} + \beta_3 \mathcal{L}_{cc}. \label{eq:cluster}
\end{align}
where $q_{ij}$ is the soft label of the embedding point $z_i$. This label measures the similarity between $z_i$ and cluster center $\lambda_j$ by Student’s t-distribution. The target distribution $p_{ij}$ is obtained by squaring $q_{ij}$ and then normalizing by the sum of the squared values across all clusters. $\beta_1$, $\beta_2$ and $\beta_3$ are three weighted factors that balance the impact of $\mathcal{L}_{zinb}$, $\mathcal{L}_{cls}$ and $\mathcal{L}_{cc}$. Their combined sum equals 1.

\section{Privacy Analysis} \label{sec:privacy_ana}

Due to the latent representation being generated in the middle of the autoencoder during inference, we just need to implement differential privacy on the encoder to achieve privacy protection according to the post-processing property of differential privacy \cite{zhu2021bias}. The privacy guarantee of DP-DCAN is proved as follows:

\begin{theorem}\label{the:privacy-loss-scCDP}
{\bf (Privacy loss of DP-DCAN)}. The privacy loss of DP-DCAN satisfies: 
\begin{align}
	\begin{split} 
		\begin{aligned}
        (\epsilon,\delta)=&(R1+ R2 +\log ((\alpha-1) / \alpha) \\
        &-(\log \delta+ \log \alpha) /(\alpha-1),\delta).
		\end{aligned}
	\end{split}
\end{align}
where $0<\delta<1$, $R1$ and $R2$ is the RDP of algorithm DPAN which is computed by Theorems~\ref{the:rdp-of-dpsgd}.
\end{theorem}

\begin{definition} \label{privacy Composition of RDP}
{\bf(Composition of RDP\cite{mironov2017renyi})}. For two randomized mechanisms $f, g$ such that $f$ is $(\alpha,R_1)$-RDP and $g$ is $(\alpha,R_2)$-RDP the composition of $f$ and $g$ which is defined as $(X, Y )$(a sequence of results), where $ X \sim f $ and $Y \sim g$, satisfies $(\alpha,R_1+R_2)-RDP$.
\end{definition}

\begin{definition}\label{definition:conversion}
{\bf (Conversion from RDP to DP~\cite{balle2020hypothesis})}. if a randomized mechanism $f : D \rightarrow \mathbb{R}$  satisfies $(\alpha,\epsilon)$-RDP ,then it satisfies$(\epsilon+\log ((\alpha-1) / \alpha)-(\log \delta+ \log \alpha) /(\alpha-1), \delta)$-DP for any $0<\delta<1$.
\end{definition}

From Definition~\ref{privacy Composition of RDP} and Definition~\ref{definition:conversion}, Theorem 1 is proofed.

\begin{theorem}\label{the:rdp-of-dpsgd} After $T$ times of iterations, the RDP of training in DPAN satisfies:
        \begin{equation}\label{equ:epsilon}
            R_{train}(\alpha)= \frac{T}{\alpha-1}\ln[\sum_{i=0}^{\alpha}\left(\begin{array}{l}
            \alpha \\ i
            \end{array}\right)(1-q)^{\alpha-i} q^{i} \exp \left(\frac{i^{2}-i}{2 \sigma^{2}}\right)].
        \end{equation}
    	where $q=\frac{L}{N}$, $\sigma$ is noise scale of training, and $\alpha > 1$ is the order.
\end{theorem}

Definitions~\ref{def:sgm} and \ref{privacy RDP privacy budget of SGM} define Sampled Gaussian Mechanism (SGM) and Rényi Differential Privacy (RDP), respectively.

\begin{definition}\label{def:sgm}
{\bf (Sampled Gaussian Mechanism (SGM)~\cite{mironov2019r})}. Let $f$ be a function mapping subsets of $S$ to $\mathbb{R}^d$. We define the Sampled Gaussian Mechanism (SGM) parameterized with the sampling rate $0 < q \leq 1$ and the  $\sigma > 0$ as
\begin{equation}
\begin{aligned}
	S G_{q, \sigma}(S) \triangleq & f(\{x: x \in S \text { is sampled with probability } q\}) \\
	&+\mathcal{N}\left(0, \sigma^{2} \mathbb{I}^{d}\right).
	\end{aligned}
\end{equation}
In DPAN, $f$ is the clipped gradient evaluation on sampled data points $f(\{x_i\}_{i\in B})$ $= \sum_{i\in B} \overline{g}_t(x_i)$. If $ \overline{g}_t$ is obtained by clipping $g_t$ with a gradient norm bound $C$, then the sensitivity of $f$ is equal to $C$.
\end{definition}

\begin{definition}\label{privacy RDP privacy budget of SGM}
{\bf(RDP privacy budget of SGM\cite{mironov2019r})}. Let $SG_{q,\sigma}$, be the Sampled Gaussian Mechanism for some function $f$. If $f$ has sensitivity 1, $SG_{q,\sigma}$ satisfies $(\alpha,R)$-RDP whenever
\begin{equation}
R \leq \frac{1}{\alpha-1} \log max(A_{\alpha}(q,\sigma),B_{\alpha}(q,\sigma)).
\end{equation}
where
\begin{equation}
\left\{\begin{array}{l}
A_{\alpha}(q, \sigma) \triangleq \mathbb{E}_{z \sim \mu_{0}}\left[\left(\mu(z) / \mu_{0}(z)\right)^{\alpha}\right], \\
B_{\alpha}(q, \sigma) \triangleq \mathbb{E}_{z \sim \mu}\left[\left(\mu_{0}(z) / \mu(z)\right)^{\alpha}\right].
\end{array}\right.
\end{equation}
with $\mu_{0} \triangleq \mathcal{N}\left(0, \sigma^{2}\right), \mu_{1} \triangleq \mathcal{N}\left(1, \sigma^{2}\right) \mbox { and } \mu \triangleq(1-q) \mu_{0}+q \mu_{1}$

Further, it holds $\forall(q,\sigma)\in(0,1], \mathbb{R}^{+*},A_{\alpha}(q,\sigma) \geq B_{\alpha}(q, \sigma) $. Thus, $ S G_{q, \sigma}$ satisfies $\left(\alpha, \frac{1}{\alpha-1} \log \left(A_{\alpha}(q, \sigma)\right)\right)$-RDP.

Finally, \cite{mironov2019r} describes a procedure to compute $A_{\alpha}(q,\sigma)$ depending on integer $\alpha$.
\begin{equation}
A_{\alpha}=\sum_{k=0}^{\alpha}\left(\begin{array}{l}.
\alpha \\ k
\end{array}\right)(1-q)^{\alpha-k} q^{k} \exp \left(\frac{k^{2}-k}{2 \sigma^{2}}\right)
\end{equation}
\end{definition}

From Definitions~\ref{def:sgm} and Definition~\ref{privacy RDP privacy budget of SGM}, the Theorem~\ref{the:rdp-of-dpsgd} is proofed.

\section{EXPERIMENTS} \label{sec:results}
Using differential privacy with partial perturbation can protect data privacy during the process of feature construction. In this section, we will explore the impact of the scope and intensity of model perturbation on model performance through a clustering task on multiple real-world scRNA-seq datasets.

\subsection{Datasets and Pre-processing}
\textit{1) Dataset:} We conduct extensive experiments on 8 real-world scRNA-seq datasets from various sequencing platforms. The detailed information is described in Table \ref{tab:datasets}. All 8 datasets are from different species, including mouse and human, as well as from different organs, such as the brain and embryo. Specifically, the numbers of cells range from 56 to 4271, and genes range from 3840 to 57241.
\begin{table}[htbp]
    \caption{The characteristics of the real scRNA-seq datasets}
    \centering
    \begin{tabular}{cccccc}
        \toprule
        \makebox[0.05\textwidth][c]{Dataset} & \makebox[0.08\textwidth][c]{Cells} & \makebox[0.08\textwidth][c]{Genes} & \makebox[0.05\textwidth][c]{Class} & \makebox[0.1\textwidth][c]{Platform} & \makebox[0.1\textwidth][c]{Reference} \\
        \midrule
        Biase & 56 & 25737 & 5 & Smart-Seq & \cite{biase2014cell} \\
        Yan & 124 & 3840 & 8 & Tang & \cite{yan2013single} \\
        Li & 561 & 57241 & 9 & SMARTer & \cite{li2017reference} \\
        Muraro & 2122 & 18915 & 9 & CEL-Seq2 & \cite{muraro2016single} \\
        Klein & 2717 & 24175 & 4 & InDrop & \cite{klein2015droplet} \\
        Romanov & 2881 & 24352 & 7 & Unknown & \cite{romanov2017molecular} \\
        Zeisel & 3005 & 19072 & 9 & Fluidigm C1 & \cite{zeisel2015cell} \\
        Zheng & 4271 & 16449 & 8 & 10X & \cite{zheng2017massively} \\
        \bottomrule
    \end{tabular}
    \label{tab:datasets}
\end{table}

\textit{2) Pre-processing:} We adopts the scRNA-seq gene expression matrix $X$ with $n$ samples as the input. First, we filter the genes that have almost no expression value since there is a large amount of technical and biological noise in the stochastic single-cell gene expression pattern. Second, size factors are calculated by dividing the median expression value of each cell by the overall dataset median. The last step is to normalize the expression matrix, ensuring a variance of 1 and adjusting the mean to 0. After normalization, 2000 highly variable genes were selected as representative genes for each cell by scanpy package.

\subsection{Implementation and Parameters Setting}
In the proposed DP-DCAN framework, the sizes of the hidden fully connected layer in the encoder are set to (256, 64), the decoder is the reverse of the encoder, and the bottleneck layer (the latent space) has a size of 32. In our two-stage training process, the optimizer for the instance stage is Adam optimizer with setting $lr=0.001$ and for the cluster stage is Adadelta with setting $lr=1.0$. We set the batch size to around 0.1 of the dataset size. To achieve an appropriate trade-off between utility and privacy, the clipping bound $C=0.1$ and $\delta=10^{-5}$. Clustering performance is measured using metrics such as Normalized Mutual Information (NMI)\cite{strehl2002cluster} and Adjusted Rand Index (ARI)\cite{hubert1985comparing}.

\subsection{Experimental Evaluation}

We evaluate the clustering performance on 8 scRNA-seq datasets in terms of both clustering methods (CL) and differential privacy algorithms (DP-ALG) shown in Table \ref{tab:result_clsmethod} and \ref{tab:result_dpmethod}. DCAN represents our proposed deep contrastive learning autoencoder network for clustering, and the combination of our proposed privacy-preserving algorithm DPAN is DP-DCAN. To explore the ability of our proposed clustering method to extract features from cells, we compare DP-DCAN with two baseline methods for single-cell clustering. scDC and bmVAE both methods use an autoencoder as a tool for feature extraction. For our proposed differential privacy method DPAN, we demonstrate its superiority by comparing it with DP-PSAC \cite{xia2023differentially}, which is a differentially private per-sample adaptive clipping algorithm based on a non-monotonic adaptive weight function. Each clustering method was run ten times to take the average. The values of the best performance metrics are bolded.

\begin{table*}[htbp]
    \caption{Performance of our method and the other clustering baseline methods on 8 scRNA-seq datasets when $\epsilon$=8.0.} 
    \label{tab:result_clsmethod}
    \centering
    \begin{tabular*}{\linewidth}{@{}c|c|c|cccccccc@{}}
        \toprule
            \makebox[0.075\textwidth][c]{Metric}& \makebox[0.14\textwidth][c]{DP-ALG} & \makebox[0.05\textwidth][c]{CL} & \makebox[0.065\textwidth][c]{Biase} & \makebox[0.065\textwidth][c]{Yan} & \makebox[0.065\textwidth][c]{Li} & \makebox[0.07\textwidth][c]{Muraro} & \makebox[0.065\textwidth][c]{Klein} & \makebox[0.095\textwidth][c]{Romanov} & \makebox[0.075\textwidth][c]{Zeisel} & \makebox[0.075\textwidth][c]{Zhang}\\
        \midrule
        \multirow{6}{*}{NMI}
         & \multirow{3}{*}{Non-private} & scDC & 96.52 & 39.91 & \textbf{92.00} & 88.71 & 90.25 & \textbf{66.62} & \textbf{76.29} & \textbf{77.23} \\
         &  & bmVAE & 61.74 & 00.00 & 82.07 & 72.81 & 60.84 & 51.50 & 61.66 & 60.17 \\
         &  & DCAN & \textbf{100.00} & \textbf{84.35} & 91.22 & \textbf{90.22} & \textbf{95.76} & 61.54 & 69.13 & 76.34 \\
        \cmidrule{2-11}
         & \multirow{3}{*}{DPAN} & scDC & 63.60 & 30.14 & 88.14 & 82.73 & \textbf{93.28} & 62.67 & 71.89 & \textbf{75.48}\\
         &  & bmVAE & 00.00 & 30.28 & 68.13 & 64.06 & 61.82 & 52.34 & 45.44 & 62.97\\
         &  & DCAN & \textbf{89.53} & \textbf{79.56} & \textbf{91.19} & \textbf{83.54} & 85.66 & \textbf{62.98} & \textbf{74.35} & 72.53\\
         
         \midrule
         \multirow{6}{*}{ARI}
         & \multirow{3}{*}{Non-private} & scDC & 98.21 & 19.66 & 80.16 & 92.55 & 86.65 & \textbf{63.75} & 72.90 & 73.72 \\
         &  & bmVAE & 34.91 & 00.00 & 69.55 & 49.06 & 37.24 & 25.22 & 32.19 & 30.43 \\
         &  & DCAN & \textbf{100.00} & \textbf{80.92} & \textbf{89.91} & \textbf{93.88} & \textbf{97.58} & 61.57 & \textbf{74.66} & \textbf{74.48} \\
        \cmidrule{2-11}
         & \multirow{3}{*}{DPAN} & scDC & 49.46 & 09.76 & 77.08 & 86.67 & \textbf{95.85} & 55.77 & 56.12 & 70.25\\
         &  & bmVAE & 00.00 & 15.61 & 53.16 & 57.22 & 51.12 & 61.32 & 44.47 & 50.42\\
         &  & DCAN & \textbf{93.40} & \textbf{68.19} & \textbf{90.27} & \textbf{88.58} & 81.18 & \textbf{68.72} & \textbf{74.17} & \textbf{70.87}\\
       \bottomrule
    \end{tabular*}
\end{table*}

\subsubsection{Clustering Methods Evaluation.}
Table \ref{tab:result_clsmethod} summarizes the clustering performance of the proposed DCAN and the baseline clustering methods on 8 scRNA-seq datasets. 
In the non-private clustering methods, DCAN achieves the best NMI and ARI on 4 and 6 of 8 scRNA-seq datasets, respectively. In the private clustering methods, DCAN achieves the best NMI and ARI on 6 and 7 of 8 scRNA-seq datasets, respectively. 
The non-private clustering methods tend to achieve better NMI and ARI on most scRNA-seq datasets than private clustering methods. This is because differential privacy algorithms introduce noise into the model preventing the model from optimizing in the true direction. 
However, appropriate noise may result in performance gains. For example, in the Li and Romanov dataset, the NMI and ARI of DCAN with DPAN are better than those without DPAN. This phenomenon could be attributed to the introduction of moderate noise, which potentially enhances the generalization performance of the model, and additional investigation is warranted to fully comprehend this effect. 
Meanwhile, we can observe that bmVAE the clustering performance is not stable. The main reason is that further simulation of the data by ZINB distribution is necessary. 
Furthermore, the clustering performance of deep embedded clustering with the contrastive learning module is better and more stable, which again proves the superiority of DCAN.
\begin{table*}[htbp]
    \caption{Performance of our method and the other differential privacy baseline method on 8 scRNA-seq datasets when $\epsilon$=8.0.} 
    \label{tab:result_dpmethod}
    \centering
    \begin{tabular*}{\linewidth}{@{}c|c|c|cccccccc@{}}
        \toprule
            \makebox[0.08\textwidth][c]{Metric}& \makebox[0.075\textwidth][c]{CL} & \makebox[0.14\textwidth][c]{DP-ALG} & \makebox[0.065\textwidth][c]{Biase} & \makebox[0.065\textwidth][c]{Yan} & \makebox[0.065\textwidth][c]{Li} & \makebox[0.075\textwidth][c]{Muraro} & \makebox[0.065\textwidth][c]{Klein} & \makebox[0.095\textwidth][c]{Romanov} & \makebox[0.075\textwidth][c]{Zeisel} & \makebox[0.075\textwidth][c]{Zhang}\\
        \midrule
        \multirow{6}{*}{NMI}
         & \multirow{2}{*}{scDC} & DP-PSAC & 32.13 & 29.33 & 36.60 & 51.53 & 61.87 & 29.69 & 45.21 & 33.46 \\
         & & DPAN & \textbf{63.60} & \textbf{30.14} & \textbf{88.14} & \textbf{82.73} &  \textbf{93.28} & \textbf{62.67} & \textbf{71.89} &  \textbf{75.48} \\
        \cmidrule{2-11}
         & \multirow{2}{*}{bmVAE} & DP-PSAC & 00.00 & 00.00 & 00.00 & 56.99 & 41.19 & 43.45 & \textbf{47.24} & 41.87 \\
         & & DPAN & 00.00 & \textbf{30.28} & \textbf{68.13} & \textbf{64.06} & \textbf{61.82} & \textbf{52.34} & 45.44 & \textbf{62.97} \\
        \cmidrule{2-11}
         & \multirow{2}{*}{DCAN} & DP-PSAC & 84.86 & 76.74 & 87.63 & 66.96 & 57.43 & 51.47 & 53.39 & 58.93 \\
         & & DPAN &  \textbf{89.53} &  \textbf{79.56} &  \textbf{91.19} &  \textbf{83.54} & \textbf{85.66} &  \textbf{62.98} &  \textbf{74.35} & \textbf{72.53} \\
         
         \midrule
         \multirow{6}{*}{ARI}
         & \multirow{2}{*}{scDC} & DP-PSAC & 23.44 & \textbf{11.98} & 16.86 & 40.08 & 59.11 & 37.95 & 33.17 & 23.90 \\
         & & DPAN & \textbf{49.46} & 09.76 & \textbf{77.08} & \textbf{86.67} &  \textbf{95.85} & \textbf{55.77} & \textbf{56.12} & \textbf{70.25} \\
         \cmidrule{2-11}
         & \multirow{2}{*}{bmVAE} & DP-PSAC & 00.00 & 00.00 & 00.00 & 54.69 & 45.66 & 58.54 & 41.23 & 34.60 \\
         & & DPAN & 00.00 & \textbf{15.61} & \textbf{53.16} & \textbf{57.22} & \textbf{51.12} & \textbf{61.32} & \textbf{44.47} & \textbf{50.42} \\
        \cmidrule{2-11}
         & \multirow{2}{*}{DCAN} & DP-PSAC & 78.70 & 62.59 & 78.28 & 70.33 & 45.40 & 61.40 & 62.21 & 47.62 \\
         & & DPAN & \textbf{93.40} &  \textbf{68.19} &  \textbf{90.27} &  \textbf{88.58} & \textbf{81.18} &  \textbf{68.72} &  \textbf{74.17} &  \textbf{70.87} \\
       \bottomrule
    \end{tabular*}
\end{table*}
\subsubsection{Differential Privacy Methods Evaluation.}
\begin{figure*}[htbp]
    \centering
	\includegraphics[width=0.8\linewidth]{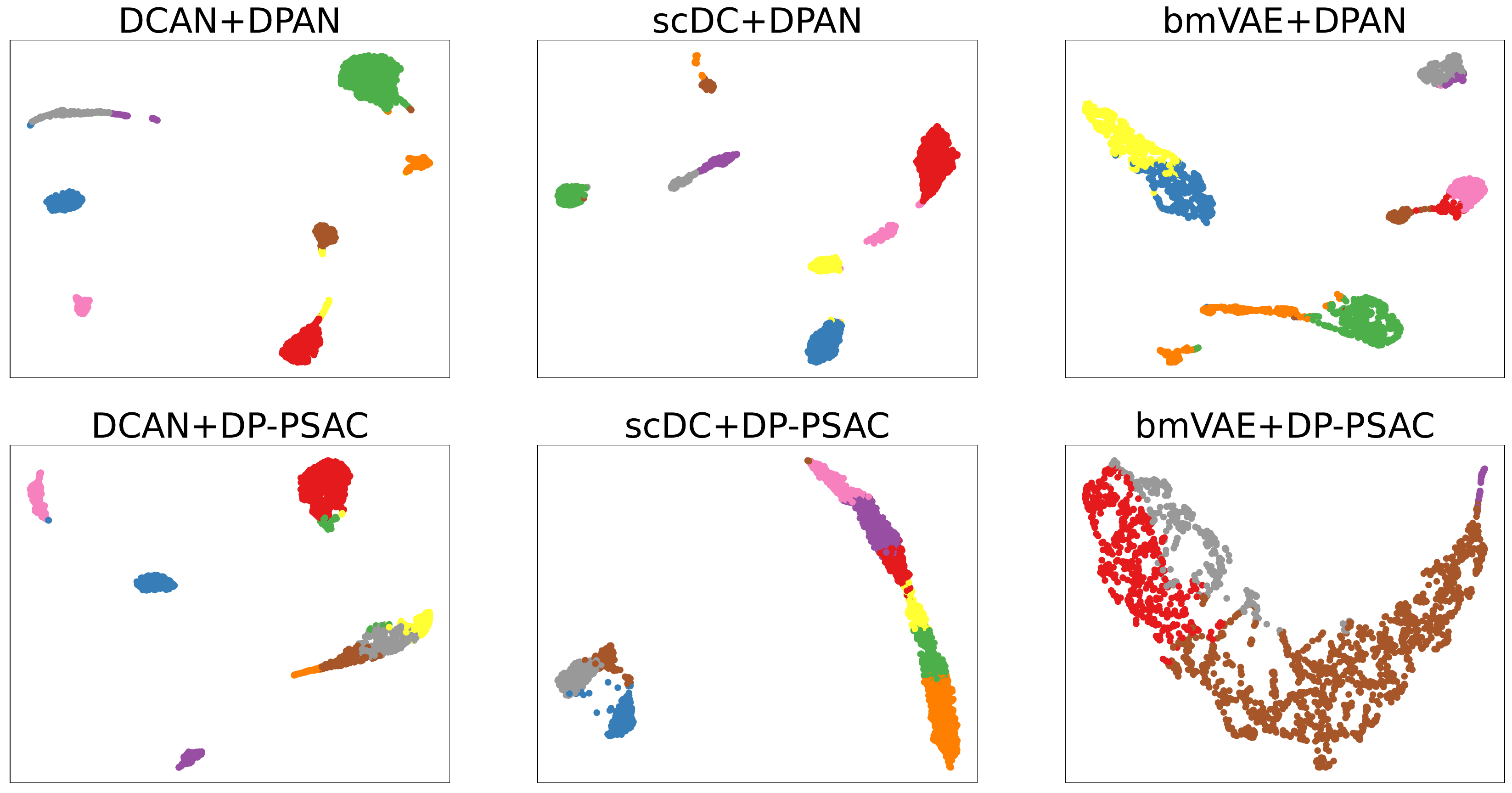}
    \caption{Comparison of clustering results with 2D visualization by UMAP on the Muraro dataset.}
    \label{fig:compare_scatter_muraro}
\end{figure*}

Table \ref{tab:result_dpmethod} shows the impact of various differential privacy algorithms on the performance of different clustering methods.
We can observe that our DPAN outperforms the other baseline differential privacy algorithm for clustering performance. 
For the 8 scRNA-seq datasets, the clustering methods with DPAN achieve the best NMI and ARI on more than 6 of these datasets, respectively. This is because the DPAN is able to mitigate the loss of performance by centrally allocating the privacy budget to the encoder part of the model. 
In clustering methods scDC and bmVAE, DPAN significantly maintains the ability of the model to learn the feature representation of the data, while DP-PSAC exhibits a tendency to excessively disrupt the model. 
Even DP-PSAC has repeatedly shown ARI and NMI values are equal 0, indicating a loss of the ability to effectively extract model features. 
In the proposed clustering method DCAN with DPAN has achieved the best performance across all datasets compared with DP-PSAC.
In summary, we can conclude that DP-DCAN performs better than the other methods under two clustering evaluation metrics. To demonstrate the intuitive discrimination ability of DP-DCAN, we employ the Uniform Manifold Approximation and Projection (UMAP)\cite{becht2019dimensionality} technique with default parameters to project the scRNA-seq data into a 2D space on the Muraro datasets in Fig. \ref{fig:compare_scatter_muraro}.

\subsection{Partial Network Perturbation vs. Entire Network Perturbation}
We conducted a comparative analysis of the proposed DP-DCAN and DPE-DCAN, the latter involving perturbation of the entire model during the training process.
\begin{figure*}[htbp]
    \centering
	\includegraphics[width=1.0\linewidth]{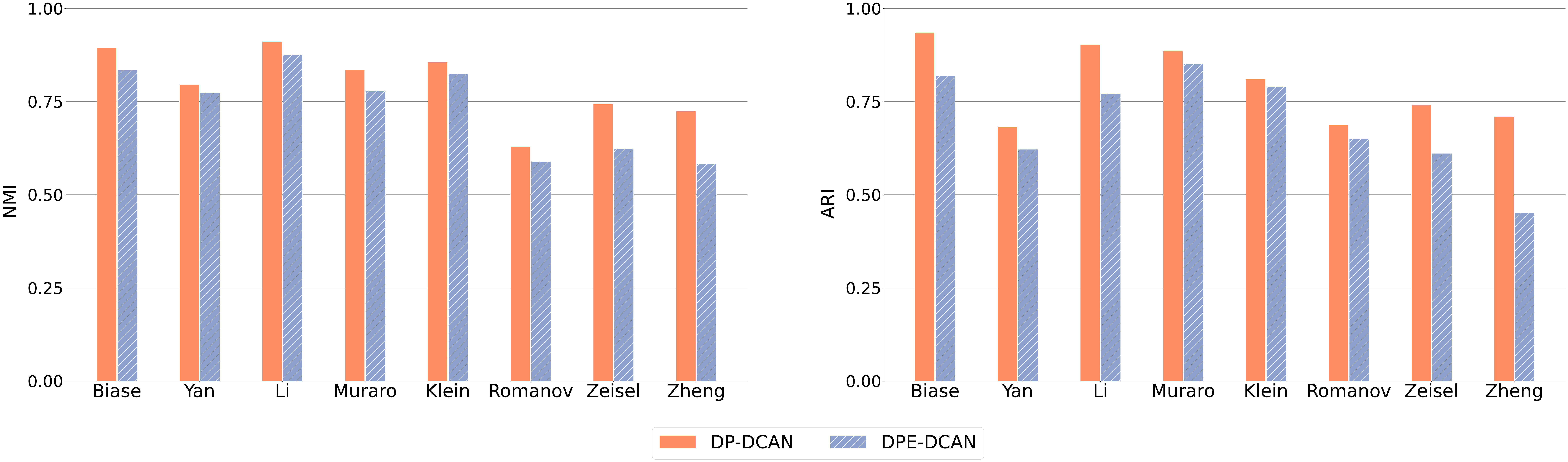}
    \caption{Clustering performance of DP-DCAN and DPE-DCAN when $\sigma=2.00$.}
    \label{fig:comp_budget_bar}
\end{figure*}

Fig. \ref{fig:comp_budget_bar} shows the clustering performance of our method against DPE-DCAN when the same noise scale $\sigma=2.00$.
For each scRNA-seq dataset, we observe that the DP-DCAN generally outperforms the DPE-DCAN. On average, the NMI and ARI scores of DP-DCAN are more than 6.32\% and 9.82\% higher than the scores of the DPE-DCAN, respectively. This is because, with the same noise scale, DP-DCAN solely introduces perturbations to the model's encoder, while DPE-DCAN necessitates perturbing the entire model. Therefore, during training, DP-DCAN demonstrates superior convergence in the model training process. Additionally, the privacy budget of DP-DCAN, with $\epsilon=8.0$, is smaller than the privacy budget of DPE-DCAN, with $\epsilon=12.0$ at the same noise intensity. 
In conclusion, this experiment provides intuitive evidence to confirm that DP-DCAN can provide reliable and remarkable performance, even with a lower privacy budget.

\begin{figure*}[htbp]
    \centering
	\includegraphics[width=1.0\linewidth]{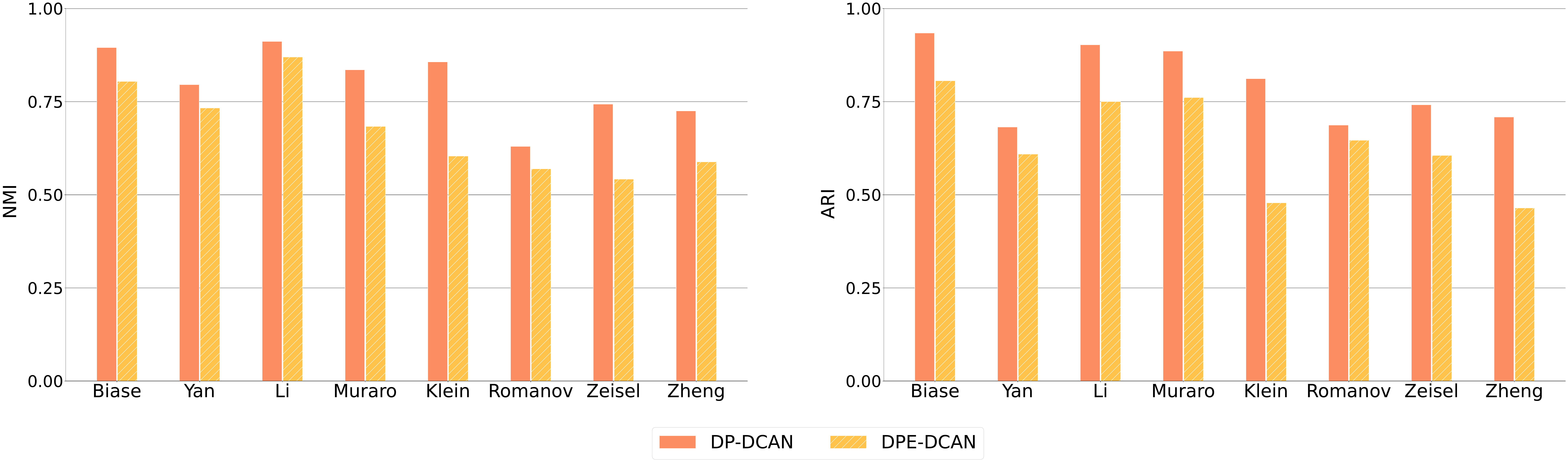}
    \caption{Clustering performance of DP-DCAN and DPE-DCAN when $\epsilon=8.0$.}
    \label{fig:comp_sigma_bar}
\end{figure*}

As depicted in Fig. \ref{fig:comp_sigma_bar}, we compare the proposed DP-DCAN and DPE-DCAN when the $\epsilon=8.0$. Obviously, DP-DCAN outperforms the DPE-DCAN for clustering performance for all scRNA-seq datasets. Even in the Klein dataset, the NMI and ARI scores of DP-DCAN are 25.25\% and 33.35\% higher than DPE-DCAN, respectively, indicating a significant performance gap. This finding underscores the importance of partial network perturbation. In the context of a designated privacy allocation, enhancing the safeguarding of the model's architecture necessitates the incorporation of increased perturbation within the gradients. Notably, the noise scale for DP-DCAN stands at 2.00, while for DPE-DCAN, it is set to 4.36. Considering the collective impact of noise intensity and the scope of noise impact, our method, DP-DCAN, employing a partially perturbed network, has exhibited exceptional efficacy across various scRNA-seq datasets.

\subsection{Unveiling Model Perturbation}
In an autoencoder network, the hidden fully connected layer can provide the functionality of feature learning and data reconstruction. To explore the impact of the scope of the perturbed layer, we apply DP-DCAN on real scRNA-seq datasets with perturbed layers from 1 to 6 with the noise scale $\sigma = 2.00$. Table \ref{tab:perturb_scope}
tabulates the average NMI and ARI values on the 8 scRNA-seq datasets for the 6 cases with DP-DCAN. As shown in Table \ref{tab:perturb_scope}, it can be clearly observed that clustering performance exhibits a fluctuating pattern with perturbed layers from 1 to 6. 
The superior performance of $layer_1$ can be ascribed to its inherent role in extracting fundamental features from the input data. Consequently, the introduction of noise to the gradients of the first layer during training may serve to augment the model's generalization capability. The diminished performance observed for $layer_{1-2}$ in comparison to $layer_1$ could potentially be attributed to the second layer’s specific task of capturing more intricate features, rendering it more susceptible to the impact of noise. The improved performance noted for $layer_{1-3}$ and $layer_{1-4}$ in contrast to $layer_{1-2}$ may be ascribed to the equilibrium achieved through the introduction of noise during the model’s training process. The decline in performance for $layer_{1-5}$ and $layer_{1-6}$ may be attributed to the broader spectrum of noise influencing these layers, consequently leading to a deterioration in overall model performance.
\begin{table*}[htbp]
	\caption{Clustering performance of DP-DCAN under different perturbation scopes.} 
    \label{tab:perturb_scope}
	\centering
    \begin{tabular}{c|cccccc}
        \toprule
        \makebox[0.12\textwidth][c]{Metric} & \makebox[0.1\textwidth][c]{$layer_1$} & \makebox[0.12\textwidth][c]{$layer_{1-2}$} & \makebox[0.12\textwidth][c]{$layer_{1-3}$} & \makebox[0.12\textwidth][c]{$layer_{1-4}$} & \makebox[0.12\textwidth][c]{$layer_{1-5}$} & \makebox[0.12\textwidth][c]{$layer_{1-6}$} \\
        \midrule
        NMI & 81.29 & 76.47 & 79.91 & 81.09 & 80.49 & 73.60 \\
        \cmidrule{1-7}
        ARI & 78.76 & 72.92 & 79.42 & 79.49 & 76.88 & 69.60 \\
         \bottomrule
    \end{tabular}
\end{table*}

\subsection{Differential Privacy Budget Analysis}
Table \ref{tab:pri_bud} shows the clustering performance of our method at different noise levels. In this experiment, we vary the privacy budget $\epsilon \in \{4.0, 6.0, 8.0\}$ on 8 scRNA-seq datasets. As depicted in Table \ref{tab:pri_bud}, we observe that as the privacy budget (privacy protection level) decreases, there is a general trend of decreasing NMI and ARI across the datasets. However, in the Biase dataset, the NMI and ARI violate this trend. Maybe this can be attributed to the introduction of moderate noise. Therefore, there is a trade-off between privacy protection and performance. This is a crucial result since smaller privacy budget values enforce stronger privacy guarantees. In cases where the privacy budget assumes larger values, such as privacy budget = 4.0, 6.0, 8.0, indicating the introduction of minimal noise into the model, the efficiencies of the model tend to converge towards heightened prediction accuracies. Balancing the privacy budget in differential privacy with model performance is a nuanced process that requires careful consideration of the privacy requirements, the sensitivity of the data, and the acceptable level of accuracy for the model's application. Moreover, DP-DCAN demonstrates strong robustness to different noise levels.

\begin{table*}[htbp]
	\caption{Clustering performance of DP-DCAN on different privacy budgets.} 
    \label{tab:pri_bud}
	\centering
    \begin{tabular*}{\linewidth}{@{}c|c|c|cccccccc@{}}
        \toprule
        \makebox[0.09\textwidth][c]{Metric} & \makebox[0.07\textwidth][c]{$\epsilon$} & \makebox[0.07\textwidth][c]{$\sigma$} & \makebox[0.08\textwidth][c]{Biase} & \makebox[0.08\textwidth][c]{Yan} & \makebox[0.08\textwidth][c]{Li} & \makebox[0.09\textwidth][c]{Muraro} & \makebox[0.09\textwidth][c]{Klein} & \makebox[0.09\textwidth][c]{Romanov} & \makebox[0.09\textwidth][c]{Zeisel} & \makebox[0.09\textwidth][c]{Zheng}\\
        \midrule
        \multirow{3}{*}{NMI}
         & 8.0 & 2.00 & 89.53 & 79.56 & 91.19 & 83.54 & 85.66 & 62.98 & 74.35 & 72.53\\
         & 6.0 & 2.49 & 96.52 & 77.49 & 90.62 & 83.47 & 85.61 & 62.13 & 74.48 & 72.53\\
         & 4.0 & 3.46 & 92.41 & 72.10 & 90.58 & 82.80 & 85.30 & 61.73 & 74.91 & 67.93\\
        \midrule
        \multirow{3}{*}{ARI}
         & 8.0 & 2.00 & 93.40 & 68.19 & 90.27 & 88.58 & 81.18 & 68.72 & 74.17 & 70.87\\
         & 6.0 & 2.49 & 98.21 & 69.42 & 89.28 & 88.41 & 81.10 & 68.23 & 75.09 & 68.12\\
         & 4.0 & 3.46 & 94.39 & 57.58 & 88.45 & 87.71 & 80.84 & 68.12 & 75.67 & 59.78\\
         \bottomrule
    \end{tabular*}
\end{table*}

\section{Conclusion} \label{sec:conclusion}

In this work, We propose DP-DCAN, a differentially private deep contrastive autoencoder network for single-cell clustering. Our scheme utilizes contrastive learning to enhance the feature extraction of the autoencoder. To protect single-cell privacy, we incorporate differential privacy for autoencoder during training. Furthermore, we design the partial network perturbation method for antoencoder. This perturbation method reduces the model dimension of the perturbation and ensures the utility. We conduct a comprehensive privacy analysis of our approach using RDP and validate our scheme through extensive experiments. The results indicate that DP-DCAN can achieve excellent performance under differential privacy protection. In future work, we will consider the possibility of partial network perturbation for neural networks other than the autoencoder, and achieve more efficient corresponding DP schemes.
%
%
%
%
\bibliography{ref}
\end{document}